\documentclass[11pt]{article}

\usepackage[final]{acl}

\usepackage{times}
\usepackage{amsmath}
\usepackage{booktabs}
\usepackage{latexsym}
\usepackage{amsfonts}
\usepackage{amssymb}
\usepackage{mathtools}

\usepackage{subcaption}

\usepackage[T1]{fontenc}
\usepackage[utf8]{inputenc}

\usepackage{microtype}

\usepackage{inconsolata}

\usepackage{graphicx}

\usepackage{multirow} 

\usepackage[dvipsnames]{xcolor}
\usepackage{tcolorbox}
\definecolor{vanillacol}{HTML}{E6F0FF}
\definecolor{cotcol}{HTML}{FFF9DB}

\newcommand{\vanillainstr}[1]{%
  \colorbox{vanillacol}{\parbox{\dimexpr\linewidth-2\fboxsep\relax}{#1}}%
}
\newcommand{\cotinstr}[1]{%
  \colorbox{cotcol}{\parbox{\dimexpr\linewidth-2\fboxsep\relax}{#1}}%
}

\newcommand{\harvard}{%
    \includegraphics[width=0.02\textwidth]{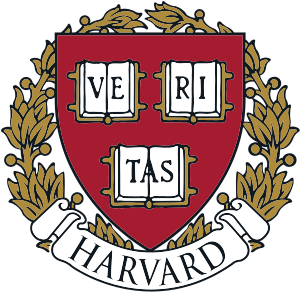} 
}

\newcommand{\squad}{\hspace{0.25em}}

\title{Large Language Models Are Bad Dice Players: LLMs Struggle to Generate Random Numbers from Statistical Distributions}
\author{
Minda Zhao \squad
Yilun Du \squad
Mengyu Wang \\
\harvard Harvard University \\
}

\begin{document}
\maketitle 
\begin{abstract}

As large language models (LLMs) transition from chat interfaces to integral components of stochastic pipelines and systems approaching general intelligence, the ability to faithfully sample from specified probability distributions has become a functional requirement rather than a theoretical curiosity. We present the first large-scale, statistically powered audit of native probabilistic sampling in frontier LLMs, benchmarking 11 models across 15 distributions. To disentangle failure modes, we employ a dual-protocol design: Batch Generation, where a model produces $N{=}1000$ samples within one response, and Independent Requests, comprising $N{=}1000$ stateless calls. We observe a sharp protocol asymmetry: batch generation achieves only modest statistical validity, with a 7\% median pass rate, while independent requests collapse almost entirely, with 10 of 11 models passing none of the distributions. Beyond this asymmetry, we reveal that sampling fidelity degrades monotonically with distributional complexity and aggravates as the sampling horizon $N$ increases. Finally, we demonstrate how the propagation of these failures into downstream real-world application tasks introduces systematic biases: models fail to enforce uniform answer-position constraints in Multiple Choice Question generation and systematically violate demographic targets in attribute-constrained text-to-image prompt synthesis. These findings indicate that current LLMs lack a functional internal sampler, necessitating external tools for applications requiring statistical guarantees.

\end{abstract}
\renewcommand{\thefootnote}{}
\footnotetext{Code and data are available at \url{https://github.com/Mininda/LLM_Bad_Dice_Player}.}
\section{Introduction}

As large language models (LLMs) transition from open-ended dialogue 
agents toward core components of complex application pipelines \cite{bubeck2023sparksartificialgeneralintelligence,bommasani2021opportunities, wang2024survey, park2023generativeagents,li2023syntheticdatagenerationlarge}, 
their capacity for statistically faithful probabilistic sampling 
has emerged as a critical functional requirement 
\cite{gu2024llmsplaydiceexploring}. The prominence of LLMs in synthetic data generation 
\cite{li2023syntheticdatagenerationlarge} has underscored the 
critical need for robust sampling mechanisms across diverse 
application scenarios \cite{shumailov2024curserecursiontraininggenerated}. For instance, in educational material generation, the need for 
faithful sampling is particularly acute. Automatic question 
generation has emerged as a promising application of LLMs, 
with the potential to reduce instructor workload and enable 
personalized learning at scale \cite{kurdi2020systematic}. 
A critical requirement in Multiple Choice Question (MCQ) construction is that correct 
answers be uniformly distributed across positions (A, B, C, D) 
to prevent test-takers from exploiting positional patterns 
\cite{haladyna2004developing}. 
Yet prior work has shown that LLMs exhibit strong positional 
preferences when selecting among options 
\cite{zheng2023judging, wang2023largelanguagemodelsfair}. 
Whether analogous biases emerge during generation, 
when models must produce MCQs adhering to uniformity constraints, 
remains unexplored. Similarly, in text-to-image generation pipelines, LLMs are increasingly employed to automatically generate diverse prompt sets \cite{hao2023optimizing, rosenman2024neuroprompts}. When constructing synthetic image datasets, practitioners often require prompts with controlled attribute distributions, such as demographic balance across gender and ethnicity, to ensure representational fairness in downstream applications \cite{sahili2024faircot}. The effectiveness of this approach hinges entirely on the LLM's ability to faithfully sample from specified distributions; if native sampling is biased, the resulting prompts will systematically deviate from target specifications. Currently, to ensure statistical rigor, the mainstream practice involves prompting LLMs to generate Python code that calls external numerical libraries such as \texttt{numpy.random} and this reliance on code-based workarounds is not incidental but systematic \cite{gao2023palprogramaidedlanguagemodels, chen2023programthoughtspromptingdisentangling, schick2023toolformerlanguagemodelsteach}. However, the ability to internalize and simulate world dynamics is increasingly viewed as a prerequisite for general intelligence \cite{lecun2022path}. Just as the community has pursued making LLMs solve mathematical 
problems without external calculators \cite{wei2022chain, lewkowycz2022solving, shao2024deepseekmath}, 
we ask whether models can generate samples from specified 
distributions without relying on external libraries. If a model must rely on an external calculator to generate even basic distributions, it suggests the model has learned 
to produce linguistic descriptions of randomness without 
acquiring the underlying functional competence 
\cite{mahowald2024dissociatinglanguagethoughtlarge}.

Recent studies have begun to examine the native sampling 
capabilities of LLMs, yielding valuable but fragmented insights. 
\citet{hopkins2023can} identified systematic biases toward 
``favorite'' numbers in random integer generation, while 
\citet{xiao2025flippingoddsreducingllm} revealed persistent 
deviations from target probabilities in coin-flip tasks. 
As the most extensive empirical study to date, 
\citet{gu2024llmsplaydiceexploring} evaluated five probability 
distributions within behavioral simulation contexts. However, their study is constrained by a sample size of $N=100$ 
insufficient for reliable convergence assessment, and an 
experimental scope limited to five simple distributions. 
Moreover, their single-prompt protocol generates all samples in one response rather than through independent calls, this design cannot determine whether models possess genuine independent sampling capabilities.

To address these gaps, we present the first large-scale, systematic evaluation of native probabilistic sampling capabilities in frontier LLMs.  Our benchmark characterizes the stochastic fidelity of 11 state-of-the-art models across a taxonomy of 15 probability distributions.  Distinguishing our work from prior small-scale studies, we evaluate each configuration at a high-resolution sample size of $N{=}1000$, enabling a statistically powered assessment of distributional convergence. Central to our methodology is a dual-protocol experimental design intended to disentangle distinct failure modes: (1) Batch Generation, where the model generates a sequence of samples within a single context window, and (2) Independent Requests, where each sample is produced via an independent call. Beyond abstract distributional benchmarks, we provide the first systematic evidence that native sampling failures carry downstream consequences: in MCQ generation, models exhibit severe positional  bias despite explicit uniformity instructions; in attribute-constrained prompt synthesis, demographic specifications are systematically violated. Our contributions are summarized as follows: (1) We demonstrate that current LLMs lack a functional internal mechanism for probabilistic sampling. (2) We reveal that sampling performance is bounded by distributional complexity. (3) We identify that increasing the sampling budget ($N$) degrades distributional adherence.

\begin{figure}[t]
    \centering
    \makebox[\columnwidth][c]{%
        \includegraphics[width=1.12\columnwidth]{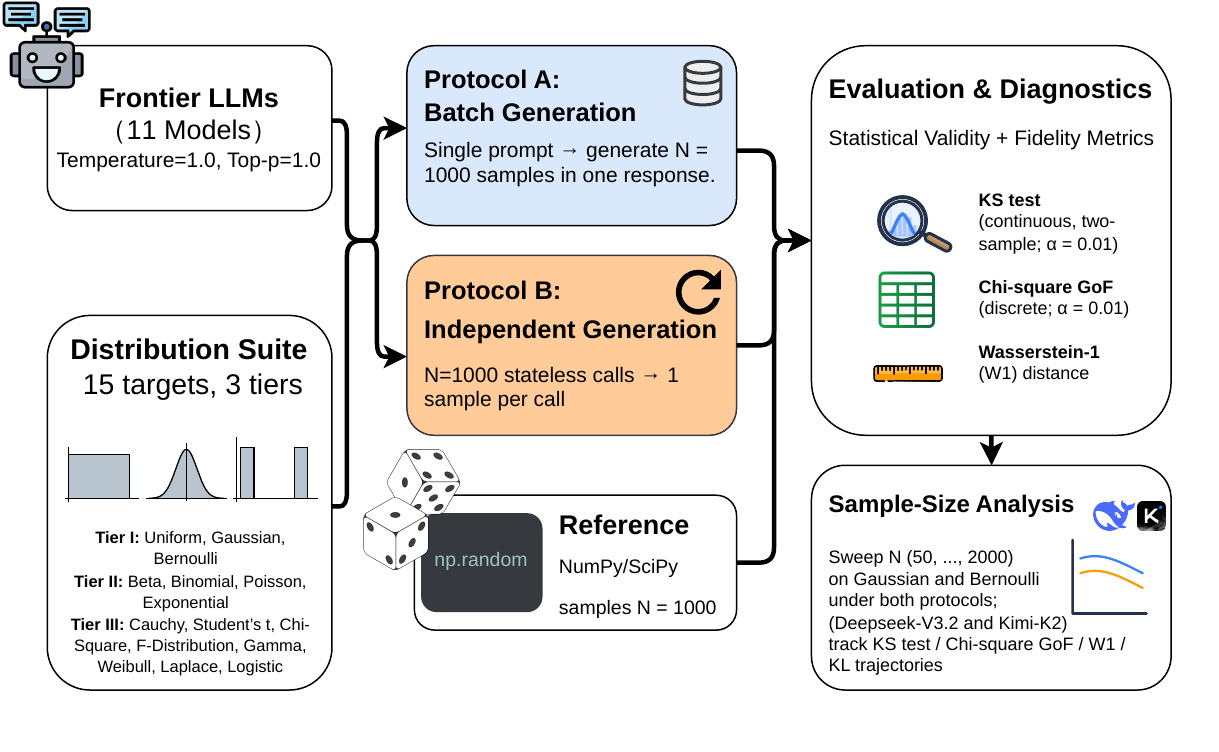}
    }
    \caption{Overview of the Evaluation Pipeline. We systematically benchmark 11 frontier LLMs across 15 probability distributions spanning three complexity tiers. The evaluation employs a dual-protocol design to disentangle failure modes: \textit{Protocol A (Batch)} produces samples sequentially within a single response, while \textit{Protocol B (Independent)} produces samples via stateless single-sample calls. Distributional fidelity is rigorously quantified using statistical validity tests (KS, $\chi^2$) and geometric metrics ($\mathcal{W}_1$) against high-precision \texttt{numpy} or \texttt{scipy} reference samples.}
    \label{fig:pipeline}
\end{figure}

\section{Related Work}

\paragraph{Sampling in LLM Applications.} LLMs now serve as core components in applications demanding statistical fidelity. In agent-based systems, \citet{park2023generativeagents}  demonstrated that LLM-powered agents can simulate believable human  behaviors in interactive environments, where behavioral diversity is  essential for realistic social dynamics. Similarly,  \citet{hao2023reasoning} showed that LLMs can  function as world simulators, predicting environment state transitions to enable multi-step planning. These paradigms rely on the model's ability to faithfully sample from complex probability spaces, as precise stochasticity is essential for maintaining behavioral diversity and modeling the inherent uncertainty of environment transitions.  Beyond agents, a broad class of generative applications
requires LLMs to produce outputs conforming to explicit distributional 
constraints. In synthetic data generation, researchers rely 
on LLMs to create diverse training sets 
\citep{li2023syntheticdatagenerationlarge, huang2025datagenunifiedsyntheticdataset}, yet 
biased sampling produces coverage gaps that propagate into downstream 
model failures \citep{shumailov2024curserecursiontraininggenerated}. 
In educational material generation, LLMs are deployed to 
automatically construct assessments, exercises, and personalized 
learning materials at scale 
\citep{kurdi2020systematic, kasneci2023chatgpt, yan2024practical}; 
applications requiring randomized test construction depend critically 
on the model's ability to honor uniformity constraints 
\citep{haladyna2004developing}. In text-to-image pipelines, 
LLMs increasingly serve as prompt generators and parsers for diffusion 
models \citep{hao2023optimizing, qin2024diffusiongpt}, producing prompt 
sets with controlled attribute distributions for dataset construction 
\citep{rosenman2024neuroprompts}; demographic balance requirements 
for fairness \citep{sahili2024faircot} hinge entirely on faithful 
sampling from target specifications.
While prior work has documented discriminative biases, positional 
preferences when LLMs select among options 
\citep{zheng2023judging, wang2023largelanguagemodelsfair}, whether 
analogous biases emerge during generation, when models must 
produce content adhering to explicit distributional constraints, 
remains unexplored. Collectively, these applications demonstrate that native sampling fidelity is not a peripheral concern but a foundational prerequisite; its limitations directly dictate the statistical integrity of downstream generative systems, necessitating a rigorous and systematic audit.

\paragraph{Empirical Studies of LLM Sampling.} A few recent studies have begun to provide preliminary evaluations of the randomness of LLM-generated outputs. Researchers have tested simple scenarios like prompting an LLM to generate uniform random bits or numbers, only to find significant deviations from true randomness \citep{hopkins2023can}. For instance, models often exhibit a favorite outcome instead of a uniform spread \citep{hopkins2023can}. Similarly,  \citet{xiao2025flippingoddsreducingllm} demonstrates a knowledge–sampling gap in Bernoulli (coin-flip) tasks: across multiple frontier LLMs (e.g., Llama-3.1, GPT-4.1-nano, DeepSeekV3, Qwen-2.5), direct sampling from ({0,1}) remains systematically biased and highly sensitive to prompt phrasing, even when the target probability is explicitly specified. Despite focusing only on these basic Uniform and Bernoulli cases, prior work already reveals that LLMs struggle to generate target distributions. These limitations underscore the lack of a large-scale, statistically rigorous benchmark capable of verifying the native sampling mechanisms of foundation models across diverse distribution types. The most comprehensive effort to date, \citet{gu2024llmsplaydiceexploring} provided a more systematic evaluation by testing five distinct probability distributions within behavioral simulation contexts. Interestingly, their findings stand in tension with earlier reports of failure, ostensibly suggesting that frontier models can approximate target distributions. However, their analysis is constrained by a limited sample size ($N{=}100$) insufficient for assessing asymptotic convergence and an experimental scope restricted to only five elementary distributions. Moreover, their reliance on batch-generation fails to disentangle sequential inter-dependencies from native sampling, leaving the question of genuine independent stochasticity unanswered. Our work addresses these gaps through the first large-scale evaluation ($N{=}1000$) across 15 distributions and 11 models, employing dual protocols to disentangle batch and independent sampling failures.

\section{Methodology}

\subsection{Problem Formulation: The Context--Fidelity Dilemma}
\label{sec:problem_formulation}

We evaluate whether an LLM can faithfully sample from a user-specified 1D target distribution $\mathcal{P}$ over $\mathbb{R}$.
Given a sampling budget $N$, the model returns samples $S_N=\{x_1,\ldots,x_N\}$, inducing an empirical measure
$\hat{\mu}_N=\frac{1}{N}\sum_{i=1}^N \delta_{x_i}$ where $\delta_x$ is the Dirac measure .
We measure fidelity by the Wasserstein-1 distance $\mathcal{W}_1(\hat{\mu}_N,\mu_{\mathcal{P}})$.
For measures on $\mathbb{R}$ with a finite first moment, $\mathcal{W}_1$ admits the CDF form \citep{vallender1974calculation}:
\begin{equation}
\mathcal{W}_1(\hat{\mu}_N,\mu_{\mathcal{P}})
=\int_{-\infty}^{\infty}\left|F_{\hat{\mu}_N}(x)-F_{\mathcal{P}}(x)\right|\,dx,
\end{equation}
where $F_{\hat{\mu}_N}$ is the empirical CDF induced by $S_N$ and $F_{\mathcal{P}}$ is the target CDF.
For an ideal i.i.d.\ sampler from $\mathcal{P}$, standard concentration results imply the expected error decreases with $N$ at the canonical $\mathcal{O}(N^{-1/2})$ rate \citep{massart1990tight,fournier2015rate}.
Our experiments show that LLMs systematically deviate from this baseline in a protocol-dependent way.

Let $\mathcal{E}(N)=\mathbb{E}\!\left[\mathcal{W}_1(\hat{\mu}_N,\mu_{\mathcal{P}})\right]$ denote the expected fidelity error.

\paragraph{Regime I: Independent Requests (stationary induced distribution).}
Under stateless calls conditioned on a fixed prompt and decoding configuration $\theta$, outputs are modeled as conditionally i.i.d.\ draws from a stationary induced distribution $\mathcal{Q}_\theta$.
Defining the intrinsic mismatch floor as $\Delta_{\mathrm{ind}} \coloneqq \mathcal{W}_1(\mu_{\mathcal{Q}_\theta}, \mu_{\mathcal{P}})$, the expected error stabilizes at:
\begin{equation}
\mathcal{E}(N) = \Delta_{\mathrm{ind}} + \mathcal{O}(N^{-1/2}) \quad \text{as } N\to\infty.
\end{equation}
The $\mathcal{O}(N^{-1/2})$ term represents sampling noise that vanishes asymptotically, leaving the irreducible bias $\Delta_{\mathrm{ind}}$.

\paragraph{Regime II: Batch Generation (Correction vs.\ Drift).}
In a single-response sequence, each $x_i$ is drawn from a history-dependent conditional $\mathcal{Q}_\theta(\cdot \mid x_{<i})$.
History dependence enables early self-correction, yet long-horizon autoregression risks accumulating deviation \citep{bengio2015scheduled,ranzato2016sequence}.
We capture this non-monotonicity through a diagnostic decomposition relative to a baseline horizon $N_0$, which denotes the smallest evaluated horizon:
\begin{equation}
\begin{aligned}
\mathrm{Corr}(N) &\coloneqq \max_{n\le N}\left( \mathcal{E}(N_0) - \mathcal{E}(n) \right), \\
\mathrm{Drift}(N) &\coloneqq \mathcal{E}(N) - \min_{n\le N} \mathcal{E}(n).
\end{aligned}
\end{equation}
This formulation yields the identity:
\begin{equation}
\mathcal{E}(N) = \mathcal{E}(N_0) - \underbrace{\mathrm{Corr}(N)}_{\text{Correction Gain}} + \underbrace{\mathrm{Drift}(N)}_{\text{Exposure Bias}}.
\end{equation}

The Context--Fidelity Dilemma arises because larger contexts can increase correction, yet beyond a critical horizon, the incremental increase in drift outweighs the incremental correction, causing net fidelity to degrade.

\subsection{Metrics}
\label{sec:metrics}

\paragraph{Wasserstein-1 Distance ($\mathcal{W}_1$).}
Serving as our primary proxy for error $\mathcal{E}$, $\mathcal{W}_1$ measures the geometric cost to transport the generated mass to the target distribution. As established in Eq.~(1), we compute the $L_1$ distance between CDFs. 

\paragraph{KL Divergence.}
We measure information loss using a histogram-based approximation to $D_{\mathrm{KL}}(\hat{p}\,\|\,p)$, where $\hat{p}$ denotes the empirical distribution of model samples and $p$ denotes the target reference distribution.

\paragraph{Statistical Validity Tests.}
To enforce a conventional binary diagnostic criterion for validity ($p>\alpha$ with $\alpha{=}0.01$), we apply distinct tests based on the support type: For continuous distributions, we employ the two-sample Kolmogorov--Smirnov (KS) test, comparing the empirical CDF of LLM-generated samples against that of high-precision reference samples ($N_{ref}=1000$) drawn from the target distribution. For discrete distributions, we employ the Chi-square goodness-of-fit test ($\chi^2$), comparing observed outcome counts against expected theoretical counts. We reject the null hypothesis (that the LLM and reference samples originate from the same underlying distribution) if the p-value falls below $\alpha=0.01$.

\section{Experiment}

\subsection{Models Under Evaluation}

We benchmark eleven frontier language models representing diverse access paradigms and architectural families to ensure comprehensive coverage of the current LLM landscape. Our selection includes \texttt{GPT-5.2}, \texttt{GPT-4o} \cite{hurst2024gpt}, \texttt{GPT-OSS-120B} \cite{agarwal2025gpt} (OpenAI), \texttt{Gemini-3-pro} \cite{comanici2025gemini}, \texttt{Gemma-3-27B} \cite{team2025gemma}  (Google), \texttt{DeepSeek-V3.2} \cite{liu2024deepseek} (DeepSeek), \texttt{Kimi-K2} \cite{team2025kimi}  (Moonshot), \texttt{Qwen3-32B} \cite{yang2025qwen3} (Alibaba), \texttt{Mistral-Small-3.2-24B}  \cite{jiang2023mistral7b} (Mistral AI), \texttt{Llama-3.3-70B} and \texttt{Llama-4-Scout-17B} \cite{grattafiori2024llama} (Meta). We use the standard default decoding settings ($T=1.0$, $\texttt{top-p}=1.0$), which are the most commonly used and provide a practical balance between randomness and output stability, making them a natural choice for evaluating native sampling ability. We further report decoding-parameter ablations in Appendix~\ref{app:Decoding}. For downstream application experiments (MCQ generation and text-to-image prompt generation), we select six representative models: \texttt{GPT-4o}, \texttt{DeepSeek-V3.2}, \texttt{Qwen3-32B}, \texttt{Llama-3.3-70B}, \texttt{Llama-4-Scout}, and \texttt{GPT-OSS-120B}.

\subsection{Distribution Sampling Evaluation}

\paragraph{Distribution Taxonomy.} To evaluate stochastic sampling behavior, we benchmark models on 15 probability distributions organized into three tiers based on entropy characteristics, support constraints, and tail behavior \citep{murphy2012machine, gelman1995bayesian} (Table~\ref{tab:distributions}). 
Tier~I includes canonical distributions such as Gaussian and Uniform, serving as standard building blocks in probabilistic modeling. 
Tier~II covers distributions with bounded supports or discrete domains (e.g., Beta, Poisson), assessing adherence to strict validity constraints. 
Tier~III comprises heavy-tailed or multi-parameter distributions (e.g., Student's $t$, Gamma), stress-testing robustness and tail-sensitive behavior beyond low-order moments. We further report a bivariate Gaussian experiment as a multivariate extension in Appendix~\ref{app:multivariate_extension}.

\begin{table}[t]
\centering
\small
\setlength{\tabcolsep}{3pt}
\begin{tabular}{@{}llll@{}}
\toprule
\textbf{Tier} & \textbf{Distribution} & \textbf{Parameters} & \textbf{Diagnostic Target} \\
\midrule
\multirow{3}{*}{\rotatebox{90}{\textbf{I}}} 
& Uniform & $a=0,\; b=1$ & Range uniformity \\
& Gaussian & $\mu=0,\; \sigma=1$ & Central tendency \\
& Bernoulli & $p=0.7$ & Binary asymmetry \\
\midrule
\multirow{4}{*}{\rotatebox{90}{\textbf{II}}} 
& Beta & $\alpha=2,\; \beta=2$ & Bounded support $[0, 1]$ \\
& Binomial & $n=10,\; p=0.5$ & Discrete counting \\
& Poisson & $\lambda=5$ & Event rate modeling \\
& Exponential & $\lambda=1$ & Positive-only support \\
\midrule
\multirow{8}{*}{\rotatebox{90}{\textbf{III}}} 
& Cauchy & $x_0=0,\; \gamma=1$ & Undefined moments \\
& Student's $t$ & $\nu=3$ & Fat tails \\
& Chi-Square & $\nu=5$ & Sum-of-squares \\
& F-Distribution & $d_1=5,\; d_2=10$ & Ratio complexity \\
& Gamma & $\alpha=2,\; \beta=2$ & Shape-scale \\
& Weibull & $k=1.5,\; \lambda=1$ & Reliability modeling \\
& Laplace & $\mu=0,\; b=1$ & Sharp peak \\
& Logistic & $\mu=0,\; s=1$ & Sigmoid symmetry \\
\bottomrule
\end{tabular}
\caption{Distribution benchmark suite with parameters and diagnostic targets.}
\label{tab:distributions}
\end{table}
\paragraph{Sampling Protocols.}
We employ two complementary protocols to disentangle distinct failure mechanisms.

\textit{Protocol A: Batch Generation.} Models receive a single prompt requesting $N{=}1000$ samples from the target distribution, generating all values within one response:
\begin{quote}
\small\textit{``You are a random number generator. Your task is to generate exactly \{N\} independent samples from a [Distribution] distribution with parameters [params].''}
\end{quote}

This protocol forces the model to condition on its generated history, probing the cumulative effect of extended context on distributional fidelity.

\textit{Protocol B: Independent Requests.} We issue $N{=}1000$ stateless calls, each requesting exactly one sample. The prompt is reduced to:
\begin{quote}
\small\textit{``Generate exactly ONE random number from a [Distribution] distribution with parameters [params]. Output ONLY the number.''}
\end{quote}

Each call is independent with no shared context, isolating the model's intrinsic priors without contextual interference. 

\paragraph{Statistical Testing and Metrics.}
For continuous distributions, we apply the two-sample Kolmogorov-Smirnov test comparing $N{=}1000$ LLM-generated samples against $N{=}1000$ reference samples from \texttt{numpy.random} and \texttt{scipy.stats}; for discrete distributions, we use Chi-square goodness-of-fit against theoretical PMFs. All tests use $\alpha = 0.01$ following \citet{gu2024llmsplaydiceexploring}, as our large sample size ($N{=}1000$) increases statistical power, requiring a stricter significance threshold to prevent the over-interpretation of minor deviations as substantive failures. A sensitivity analysis with alternative significance thresholds is provided in Appendix~\ref{app:alpha_sensitivity}. We additionally report Wasserstein-1 distance ($\mathcal{W}_1$) and KL divergence for fine-grained fidelity quantification.

\paragraph{Sample-Size Scaling Analysis.}
To characterize convergence trajectories and identify collapse thresholds, we sweep sample sizes $N \in \{50, 100, 200, 300, \ldots, 1000, 1500, 2000\}$ for Gaussian and Bernoulli sampling with DeepSeek-V3.2 and Kimi-K2 under both batch and independent protocols. At each checkpoint, we report $\mathcal{W}_1$, KL divergence, and the corresponding goodness-of-fit statistic: Kolmogorov-Smirnov (KS) for Gaussian and $\chi^2$ for Bernoulli.

\subsection{Downstream Applications}

\paragraph{MCQ Generation.}
To examine whether sampling deficiencies propagate to structured generation, we design an MCQ benchmark requiring models to produce $N{=}1000$ medical multiple-choice questions via independent calls. Crucially, prompts explicitly instruct that the position of the correct answer should be randomly and uniformly distributed among A, B, C, D to ensure no positional bias. We extract designated correct answers' position from each generated question and perform $\chi^2$ goodness-of-fit tests against the uniform target (25\% per option). This directly tests whether models can internalize uniformity constraints during content creation.

\paragraph{Attribute-Constrained Prompt Generation.}
We further stress-test native sampling in a semantically grounded setting where distributional constraints are entangled with natural language generation. Each model generates $N{=}1000$ text-to-image prompts via independent calls, each describing a person wearing a coat. Four attributes must independently conform to prescribed target distributions: Gender (Male 49.5\%, Female 50.5\%) and Race/Ethnicity (White 57.5\%, Hispanic 20.0\%, Black 12.6\%, Asian 6.5\%, Other 3.4\%) derived from \citet{uscensus_acs_2024}; Height following $\mathcal{N}(169, 10^2)$ cm; and Coat Color uniformly distributed over seven categories. We apply $\chi^2$ tests for categorical attributes and KS tests for height. This task evaluates whether LLMs can faithfully sample from explicit distributional specifications when probability constraints must be realized through semantically coherent text.

\section{Results}
\label{sec:result}

\subsection{Distribution Sampling}

\paragraph{Protocol-Dependent Sampling Fidelity.}
Table~\ref{tab:w1_results_batch_generation} shows that batch generation achieves modest statistical validity: the leading model passes 40\% of distributions, while the median pass rate is 7\%. In stark contrast, Table~\ref{tab:w1_results_independent_requests} shows a near-complete failure under independent sampling, with 10 of the 11 models failing to pass any distribution. This protocol asymmetry is not attributable to distributional difficulty: examining Uniform (the simplest benchmark), Wasserstein distances amplify from $\mathcal{W}_1{\approx}0.01$ (batch) to $\mathcal{W}_1{\approx}0.15$ (independent) across models. The stark contrast in performance suggests that valid sampling depends critically on long-context dependencies, rather than being reliably supported by isolated, stateless sampling.

\begin{table}[t]
\centering
\scriptsize
\setlength{\tabcolsep}{0.5pt}
\begin{tabular}{l|ccccccccccc}
\toprule
\textbf{Dist.} & \rotatebox{60}{\tiny \textbf{GPT-5.2}} & \rotatebox{60}{\tiny \textbf{Gemini-3}} & \rotatebox{60}{\tiny \textbf{GPT-4o}} & \rotatebox{60}{\tiny \textbf{DeepSeek}} & \rotatebox{60}{\tiny \textbf{Qwen3}} & \rotatebox{60}{\tiny \textbf{Gemma-3}} & \rotatebox{60}{\tiny \textbf{Mistral-3.2}} & \rotatebox{60}{\tiny \textbf{Kimi-K2}} & \rotatebox{60}{\tiny \textbf{Llama-3.3}} & \rotatebox{60}{\tiny \textbf{Llama-4}} & \rotatebox{60}{\tiny \textbf{GPT-OSS}} \\
\midrule
\multicolumn{12}{l}{\textit{Discrete Distributions}} \\
Bernoulli & 0.08 & 0.04 & \textbf{4e-05}$^*$ & 0.02 & 0.12 & 0.06 & 0.07 & \textbf{0.03}$^*$ & 0.16 & 0.13 & \textbf{0.02}$^*$ \\
Binomial & 0.59 & 0.19 & 0.26 & 0.20 & 0.80 & 1.1 & 0.53 & 0.81 & 0.32 & 0.56 & 0.58 \\
Poisson & 0.64 & 0.32 & 0.26 & 0.21 & 1.5 & 1.0 & 0.95 & 0.62 & 0.60 & 0.62 & 0.71 \\
\midrule
\multicolumn{12}{l}{\textit{Continuous Distributions}} \\
Uniform & \textbf{0.02}$^*$ & \textbf{0.01}$^*$ & \textbf{0.02}$^*$ & \textbf{9e-03}$^*$ & 0.10 & \textbf{0.03}$^*$ & 0.15 & 0.03 & 0.07 & 0.19 & \textbf{0.03}$^*$ \\
Gaussian & \textbf{0.13}$^*$ & 0.15 & \textbf{0.10}$^*$ & 0.43 & 0.21 & 0.15 & 0.31 & 0.17 & 0.36 & 0.26 & 0.23 \\
Beta & 0.08 & 0.06 & 0.10 & 0.06 & 0.19 & 0.06 & 0.06 & \textbf{0.03}$^*$ & 0.08 & 0.10 & 0.08 \\
Exp & 0.43 & 0.11 & 0.24 & 0.30 & 0.21 & 0.38 & 0.86 & 0.52 & 0.25 & 0.35 & 0.42 \\
Cauchy & 5.4 & 5.0 & \textbf{3.3}$^*$ & 5.9 & 5.4 & 6.6 & 6.1 & 5.7 & 6.4 & 6.0 & 6.0 \\
$t$ & 0.75 & 0.46 & \textbf{0.13}$^*$ & 0.48 & 0.72 & 0.48 & 0.43 & \textbf{0.26}$^*$ & 0.37 & 0.50 & 0.52 \\
$\chi^2$ & 0.75 & 1.0 & 0.93 & 1.2 & 1.5 & 1.6 & 4.5 & 1.8 & 0.91 & \textbf{0.66}$^*$ & 0.59 \\
$F$ & 0.43 & 0.20 & 0.45 & 0.14 & 0.40 & 0.98 & 0.71 & 0.79 & 0.42 & 0.50 & 0.90 \\
Gamma & 0.33 & 1.0 & 0.78 & 0.84 & 1.3 & 2.0 & 1.5 & 1.5 & 2.4 & 1.8 & 1.7 \\
Weibull & 0.29 & 0.13 & 0.32 & 0.36 & 0.34 & 0.13 & 0.37 & 0.51 & 0.67 & 0.33 & 0.33 \\
Laplace & 0.28 & \textbf{0.15}$^*$ & \textbf{0.32}$^*$ & 0.35 & 1.2 & 0.35 & 0.69 & 0.44 & 0.46 & 0.50 & 0.74 \\
Logistic & 0.99 & 0.37 & 0.73 & 0.54 & 2.1 & 0.40 & 0.51 & 0.39 & 1.7 & 0.92 & 0.52 \\
\midrule
\textbf{Pass Rate} & \textbf{13\%} & \textbf{13\%} & \textbf{40\%} & \textbf{7\%} & 0\% & \textbf{7\%} & 0\% & \textbf{20\%} & 0\% & \textbf{7\%} & \textbf{13\%} \\
\bottomrule
\end{tabular}
\caption{Wasserstein Distance $\mathcal{W}_1$ (Batch Generation).
Lower $\mathcal{W}_1$ indicates better distributional fit.
$^*$ denotes passing the statistical test (p $> 0.01$): $\chi^2$ GoF for discrete, two-sample KS for continuous.}
\label{tab:w1_results_batch_generation}
\end{table}

\begin{table}[t]
\centering
\scriptsize
\setlength{\tabcolsep}{1pt}
\begin{tabular}{l|ccccccccccc}
\toprule
\textbf{Dist.} & \rotatebox{60}{\tiny \textbf{GPT-5.2}} & \rotatebox{60}{\tiny \textbf{Gemini-3}} & \rotatebox{60}{\tiny \textbf{GPT-4o}} & \rotatebox{60}{\tiny \textbf{DeepSeek}} & \rotatebox{60}{\tiny \textbf{Qwen3}} & \rotatebox{60}{\tiny \textbf{Gemma-3}} & \rotatebox{60}{\tiny \textbf{Mistral}} & \rotatebox{60}{\tiny \textbf{Kimi-K2}} & \rotatebox{60}{\tiny \textbf{Llama-3.3}} & \rotatebox{60}{\tiny \textbf{Llama-4}} & \rotatebox{60}{\tiny \textbf{GPT-OSS}} \\
\midrule
\multicolumn{12}{l}{\textit{Discrete Distributions}} \\
Bernoulli & 0.32 & 0.32 & 0.31 & 0.12 & 0.29 & 0.32 & 0.27 & 0.18 & 0.32 & \textbf{0.02}$^*$ & 0.32 \\
Binomial & 1.2 & 0.80 & 0.83 & 1.2 & 5.0 & 1.4 & 1.5 & 0.87 & 1.4 & 1.2 & 1.0 \\
Poisson & 1.0 & 0.98 & 0.52 & 0.99 & 1.7 & 2.5 & 0.82 & 0.71 & 2.3 & 1.1 & 1.5 \\
\midrule
\multicolumn{12}{l}{\textit{Continuous Distributions}} \\
Uniform & 0.15 & 0.20 & 0.16 & 0.15 & 0.51 & 0.17 & 0.17 & 0.17 & 0.34 & 0.28 & 0.16 \\
Gaussian & 0.57 & 0.70 & 0.27 & 0.49 & 0.82 & 0.72 & 0.40 & 0.55 & 0.44 & 0.83 & 0.54 \\
Beta & 0.13 & 0.14 & 0.08 & 0.11 & 1.5 & 0.15 & 0.10 & 0.16 & 0.17 & 0.11 & 0.13 \\
Exp & 0.61 & 0.50 & 0.27 & 0.19 & 0.71 & 0.49 & 0.61 & 0.40 & 0.52 & 0.44 & 0.45 \\
Cauchy & 3.1 & 3.2 & 2.8 & 8.3 & 3.7 & 3.6 & 3.0 & 3.0 & 3.7 & 3.8 & 2.7 \\
$t$ & 0.59 & 0.56 & 0.40 & 0.95 & 1.1 & 1.4 & 0.60 & 0.39 & 1.1 & 1.2 & 0.68 \\
$\chi^2$  & 1.7 & 1.7 & 1.1 & 11.6 & 2.5 & 2.9 & 1.6 & 1.1 & 2.1 & 3.1 & 1.8 \\
$F$ & 0.54 & 0.56 & 0.56 & 3.7 & 3.5 & 1.2 & 0.49 & 0.32 & 0.65 & 0.31 & 0.57 \\
Gamma & 1.1 & 1.7 & 1.2 & 4.8 & 2.4 & 1.8 & 1.1 & 1.5 & 1.7 & 2.1 & 1.4 \\
Weibull & 0.35 & 0.40 & 0.30 & 2.3 & 0.75 & 0.47 & 0.50 & 0.26 & 0.40 & 0.31 & 0.26 \\
Laplace & 0.61 & 0.71 & 0.36 & 0.95 & 1.0 & 0.64 & 0.65 & 0.49 & 1.1 & 1.0 & 0.51 \\
Logistic & 0.88 & 0.80 & 0.66 & 0.75 & 1.3 & 1.2 & 0.91 & 0.90 & 1.3 & 1.4 & 0.60 \\
\midrule
\textbf{Pass Rate} & 0\% & 0\% & 0\% & 0\% & 0\% & 0\% & 0\% & 0\% & 0\% & \textbf{7\%} & 0\% \\
\bottomrule
\end{tabular}
\caption{Wasserstein Distance $\mathcal{W}_1$ (Independent Requests).
Lower $\mathcal{W}_1$ indicates better distributional fit.
$^*$ denotes passing the statistical test (p $> 0.01$): $\chi^2$ GoF for discrete, two-sample KS for continuous.}
\label{tab:w1_results_independent_requests}
\end{table}

\begin{figure}[t]
\centering
\includegraphics[width=\columnwidth]{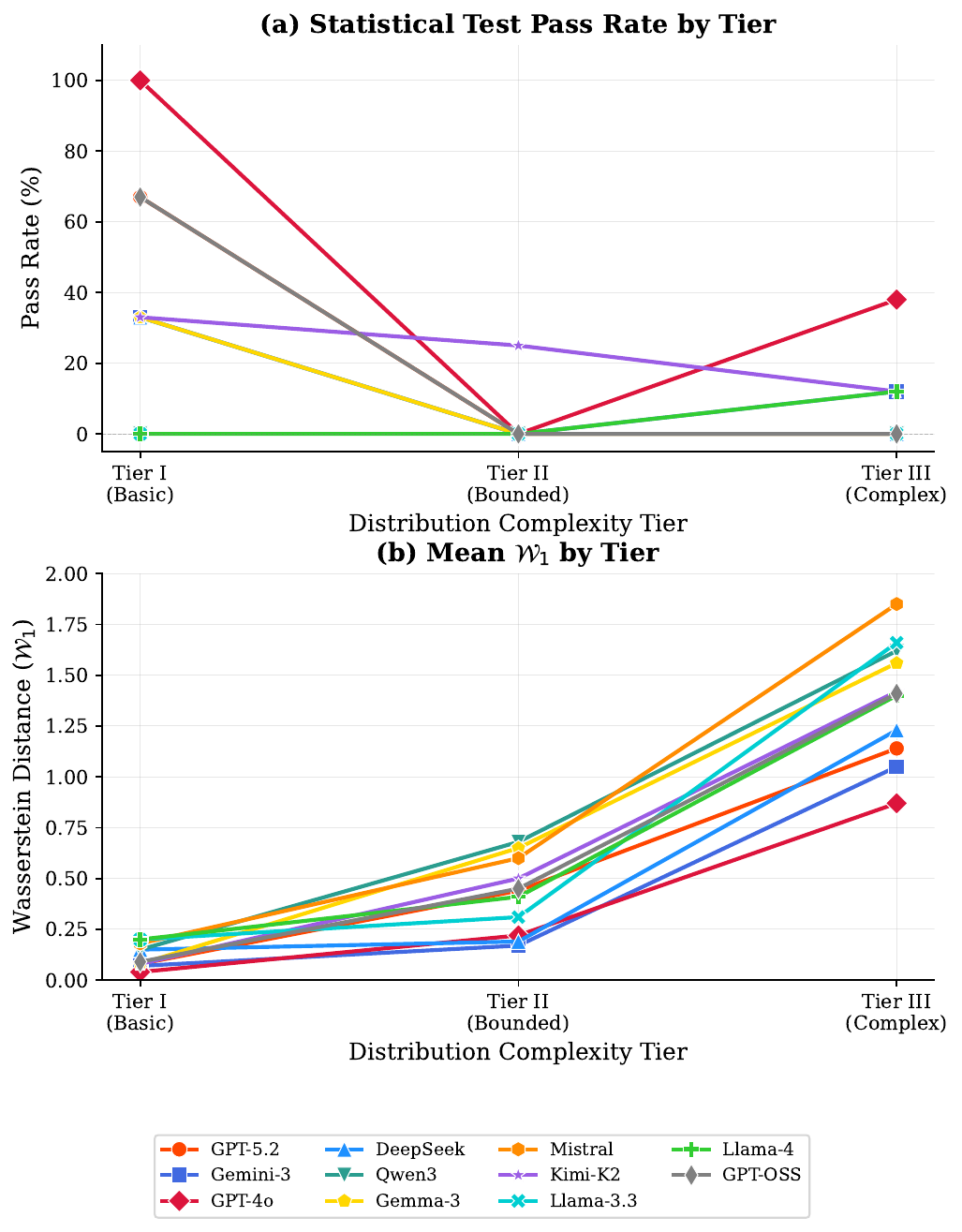}
\caption{Distribution Complexity vs.\ Sampling Fidelity.
(a) Statistical test pass rate decreases as distribution complexity increases from Tier I (Fundamental Priors) to Tier III (Heavy-Tailed \& Complex).
(b) Mean Wasserstein distance $\mathcal{W}_1$ increases with complexity, indicating poorer distributional fit.}
\label{fig:tier_comparison}
\end{figure}

\paragraph{Complexity Stratification.}
We further stratify distributions by complexity tier to examine how sampling fidelity varies with increasing distribution complexity. Figure~\ref{fig:tier_comparison} visualizes this dual trend: panel (a) shows pass rates declining monotonically across tiers for most models, with GPT-4o exhibiting steepest degradation from perfect Tier~I performance; panel (b) demonstrates the inverse relationship, $\mathcal{W}_1$ distances rise systematically with tier complexity, diverging from ${\sim}0.1$ (Tier~I) to ${\sim}1.5$ (Tier~III) across models. This coupled pattern, with declining validity alongside escalating distributional distance, shows that increasing structural constraints are associated with progressively worse performance across distribution tiers.

\begin{figure}[t]
\centering
\includegraphics[width=1.05\linewidth]{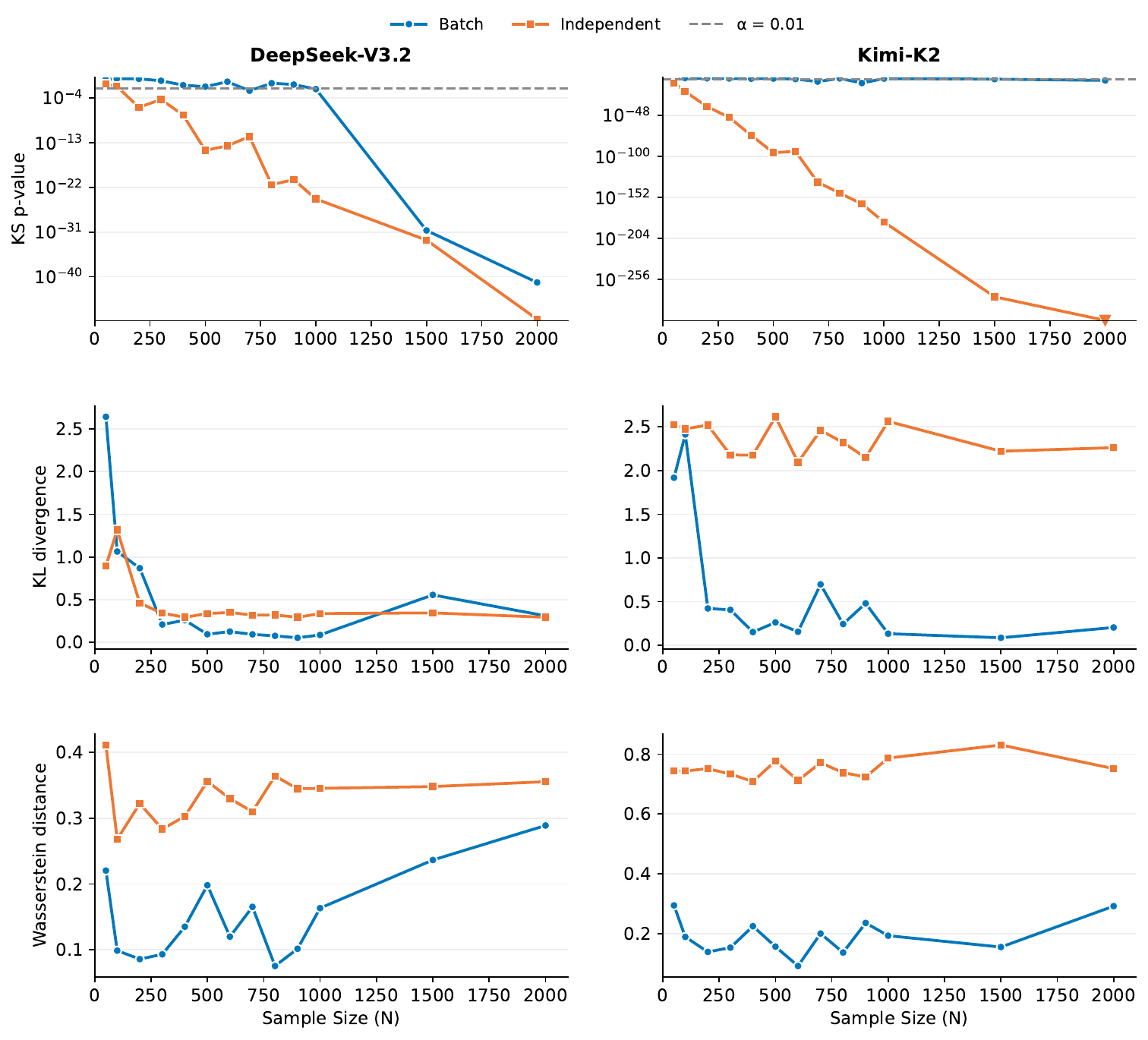}
\caption{Effect of sample size ($N$) on the Gaussian sampling quality of DeepSeek-V3.2 and Kimi-K2. The dashed line indicates the Kolmogorov-Smirnov (KS) test significance threshold at $\alpha=0.01$. Downward triangles in the Kimi-K2 KS p-value panel indicate p-values below the double-precision normal limit (2.23 × $10^{-308}$.)}
\label{fig:sample_size_effect_Gaussian}
\end{figure}

\begin{figure}[t]
\centering
\includegraphics[width=1.05\linewidth]{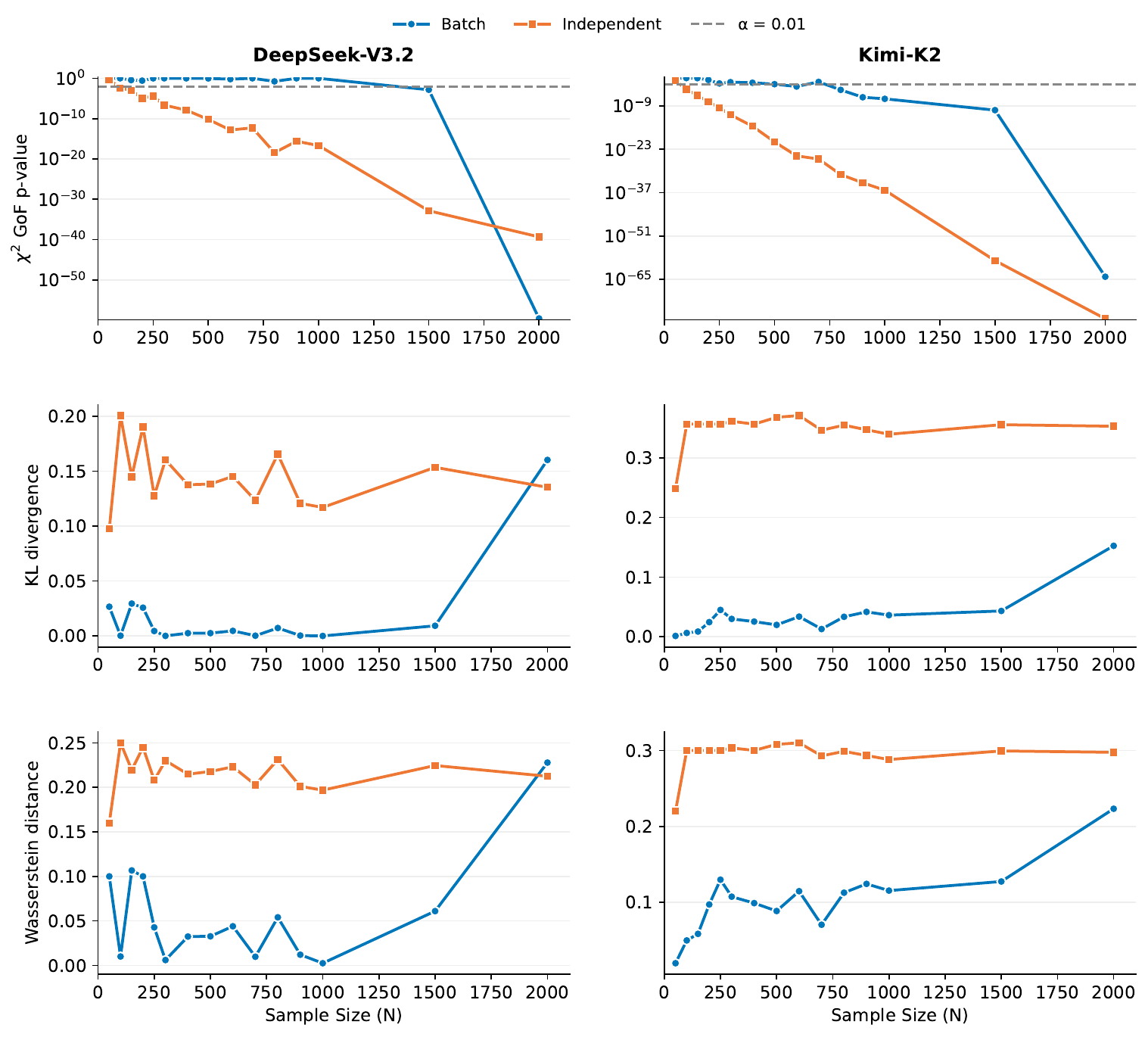}
\caption{Effect of sample size ($N$) on the Bernoulli sampling quality of DeepSeek-V3.2 and Kimi-K2. The dashed line indicates the $\chi^2$ goodness-of-fit test significance threshold at $\alpha=0.01$.}
\label{fig:sample_size_effect_Bernoulli}
\end{figure}

\paragraph{Sample-Size Trajectories.}
To examine how sampling budget affects distributional adherence, we analyze Gaussian and Bernoulli generation trajectories for $N \in \{50, \ldots, 2000\}$ (Figure~\ref{fig:sample_size_effect_Gaussian} and ~\ref{fig:sample_size_effect_Bernoulli}). The results reveal a distinct inverse scaling trend. While initial fluctuations in KL and $\mathcal{W}_1$ are attributable to finite-sample variance, the long-run behavior contradicts standard convergence expectations. Batch generation exhibits pronounced degradation: as $N$ exceeds 1000, $\mathcal{W}_1$ distances rise steadily, coinciding with KS p-values collapsing below the significance threshold ($\alpha=0.01$). Although Independent requests fail the KS test across all $N$, they display a parallel drift, with $\mathcal{W}_1$ increasing gradually. These trajectories confirm that for current LLMs, larger sample sizes reveal the statistically significant discrepancy that is invisible at small $N$.

\subsection{Downstream Applications}

\paragraph{MCQ Generation: Positional Bias Persists Despite Explicit Instructions.}
Table~\ref{tab:mcq_distribution} quantifies generative positional bias in MCQ construction. Despite explicit prompts requiring uniform distribution of correct answers across A/B/C/D positions, all six models exhibit severe and statistically significant bias ($p < 0.001$). GPT-OSS-120B shows the most extreme skew, placing 54.6\% of correct answers at position C and only 4.5\% at position A. GPT-4o favors position B (46.8\%) while nearly ignoring D (5.5\%). Notably, no model approaches the uniform 25\% target for any position. These results demonstrate that sampling deficiencies are not confined to abstract numerical generation but propagate directly into structured content creation, fundamentally compromising the reliability of LLM-generated evaluation materials.

\begin{table}[t]
\centering
\footnotesize
\setlength{\tabcolsep}{3pt}
\begin{tabular}{@{}lccccccc@{}}
\toprule
Model & A (\%) & B (\%) & C (\%) & D (\%) & $\chi^2$ & p \\
\midrule
GPT-4o & 12.6 & 46.8 & 35.1 & 5.5 & 444.5 & $<$.001 \\
Llama-3.3-70B & 17.2 & 32.9 & 42.9 & 7.0 & 307.1 & $<$.001 \\
DeepSeek-V3.2 & 16.9 & 28.2 & 36.8 & 18.1 & 105.1 & $<$.001 \\
Qwen3-32B & 21.8 & 35.6 & 31.5 & 11.1 & 143.2 & $<$.001 \\
Llama-4-Scout & 28.6 & 42.3 & 22.7 & 6.4 & 265.4 & $<$.001 \\
GPT-OSS-120B & 4.5 & 27.7 & 54.6 & 13.2 & 577.2 & $<$.001 \\
\midrule
Uniform & 25.0 & 25.0 & 25.0 & 25.0 & -- & -- \\
\bottomrule
\end{tabular}
\caption{MCQ Answer Distribution Bias (English, Temperature=1.0, N=1000). Target: Uniform 25\% per option. All models show significant bias (p $<$ 0.001).}
\label{tab:mcq_distribution}
\end{table}

\paragraph{Attribute-Constrained Prompt Generation: Systematic Distributional Violations.}
Table~\ref{tab:joint_results} reveals pervasive failures when models must translate explicit distributional specifications into semantically coherent text. For Gender, models exhibit opposing biases: GPT-4o overrepresents males (75.0\% vs.\ target 49.5\%), while Llama-4 drastically overrepresents females (97.2\%). For Race/Ethnicity, models systematically over-sample certain groups (GPT-4o: 33.5\% Asian vs.\ target 6.5\%) while severely under-representing others (GPT-4o: 0\% Hispanic vs.\ target 20.0\%; Other category: 0\% across four of six models vs.\ target 3.4\%). The Height distribution reveals variance collapse: all models produce $\sigma \approx 1$--6 cm versus the target $\sigma = 10$ cm, with KS statistics exceeding 0.37 across all models. For Coat Color, models collapse onto modal preferences, with Llama-3.3 generating 96\% green coats and GPT-OSS favoring red (54\%), completely ignoring the uniform specification. These failures persist despite prompts containing precise numerical targets, confirming that LLMs cannot internalize distributional constraints when sampling must occur through natural language generation rather than raw numerical output.

\begin{table}[t]
\centering
\scriptsize
\setlength{\tabcolsep}{1pt}

\begin{tabular}{@{}lcccccc|c@{}}
\toprule
& \rotatebox{45}{GPT-4o} & \rotatebox{45}{DeepSeek} & \rotatebox{45}{Qwen3} & \rotatebox{45}{Llama-3.3} & \rotatebox{45}{Llama-4} & \rotatebox{45}{GPT-OSS} & \rotatebox{45}{Target} \\
\midrule
\textbf{Task: Gender} \\
Male (\%) & 75.0 & 60.4 & 29.3 & 31.7 & 2.8 & 21.3 & 49.5 \\
Female (\%) & 25.0 & 39.6 & 70.7 & 68.3 & 97.2 & 78.7 & 50.5 \\
$\chi^2$ & 260.3 & 47.6 & 163.1 & 126.6 & 872.1 & 317.9 & -- \\
\midrule
\textbf{Task: Race/Ethnicity} \\
White (\%) & 41.8 & 54.5 & 57.2 & 70.5 & 51.3 & 45.0 & 57.5 \\
Hispanic (\%) & 0.0 & 6.0 & 27.1 & 6.5 & 42.9 & 10.3 & 20.0 \\
Black (\%) & 24.7 & 10.7 & 10.6 & 6.0 & 1.1 & 15.6 & 12.6 \\
Asian (\%) & 33.5 & 28.5 & 4.9 & 17.0 & 4.7 & 29.1 & 6.5 \\
Other (\%) & 0.0 & 0.3 & 0.2 & 0.0 & 0.0 & 0.0 & 3.4 \\
\midrule
\textbf{Task: Height} \\
$\mu$ (cm) & 172.7 & 171.3 & 171.9 & 172.5 & 172.5 & 173.0 & 169.0 \\
$\sigma$ (cm) & 3.0 & 5.6 & 3.9 & 0.9 & 1.5 & 3.9 & 10.0 \\
KS stat & 0.51 & 0.37 & 0.39 & 0.61 & 0.53 & 0.52 & -- \\
\midrule
\textbf{Task: Coat Color} \\
Black (\%) & 1 & 8 & 2 & 0 & 4 & 0 & 14.3 \\
White (\%) & 3 & 15 & 2 & 0 & 8 & 0 & 14.3 \\
Red (\%) & 28 & 23 & 39 & 0 & 1 & 54 & 14.3 \\
Blue (\%) & 0 & 1 & 20 & 0 & 19 & 2 & 14.3 \\
Green (\%) & 54 & 28 & 28 & 96 & 55 & 29 & 14.3 \\
Yellow (\%) & 11 & 13 & 4 & 0 & 7 & 14 & 14.3 \\
Brown (\%) & 4 & 11 & 5 & 2 & 7 & 1 & 14.3 \\
\bottomrule
\end{tabular}
\caption{Distribution fidelity in text-to-image prompt generation ($N{=}1000$). Models are prompted to sample attributes according to U.S. Census targets (Gender, Race) and standard statistical forms (Height, Color). All deviations are significant ($p < 0.001$).}
\label{tab:joint_results}
\end{table}

\section{Discussion}

\paragraph{LLMs Lack a Functional Internal Sampler.}
Our results provide compelling evidence that current LLMs do not possess a genuine internal mechanism for probabilistic sampling. The most striking finding is the near-total failure under independent sampling: 10 of 11 models achieve exactly 0\% pass rate when generating samples without shared context. This stands in sharp contrast to batch generation, where models achieve modest pass rates through within-context self-correction. The implication is clear: what appears to be sampling capability in batch mode is in fact an emergent property of autoregressive conditioning, not an internalized understanding of probability distributions. When this contextual scaffolding is removed, models default to systematic internal biases that produce statistically invalid outputs. The apparent stochasticity of LLM outputs is therefore not grounded in distributional competence.

\paragraph{Complexity Amplifies Failure.}
Figure~\ref{fig:tier_comparison} reveals a consistent relationship between distribution complexity and sampling failure. Pass rates decline from Tier~I to Tier~III, while mean $\mathcal{W}_1$ increases correspondingly. Heavy-tailed distributions such as Cauchy and Chi-Square prove particularly challenging, with no model passing the statistical tests. This pattern suggests that LLMs can only approximate distributional forms that are well-represented in their training corpora. When confronted with mathematically complex distributions requiring precise handling of bounded supports, undefined moments, or multi-parameter dependencies, models fail to generalize beyond superficial pattern matching. The fidelity gap between simple and complex distributions underscores a fundamental limitation: LLMs learn to mimic the surface statistics of familiar distributions without acquiring the underlying mathematical structure.

\paragraph{Inverse Scaling Under Increasing Sampling Budget.}
Increasing sample size should improve distributional convergence. Instead, contrary to asymptotic convergence, distributional fidelity degrades as the requested horizon $N$ grows. Figure~\ref{fig:sample_size_effect_Gaussian} and \ref{fig:sample_size_effect_Bernoulli} reveal a consistent inverse-scaling signature across diagnostics. In batch generation, $\mathcal{W}_1$ exhibits a clear regime shift: after an early improvement at short horizons, it turns upward and increases with $N$, consistent with length-amplified degradation in long sequences. Crucially, this is not a batch-only artifact. Under independent requests, the model is already invalid at small $N$ (KS p-values below threshold), yet $\mathcal{W}_1$ still drifts upward with $N$, indicating that larger budgets expose progressively larger geometric mismatch even without shared context. As $N$ increases, the accumulated discrepancy becomes statistically undeniable, driving KS p-values to vanishing levels. These findings reveal that expanding the sample budget unmasks fundamental distributional mismatches that remain statistically latent in smaller samples, particularly within batch generation regimes.

\section{Limitations}

Our findings are empirical rather than theoretical: they demonstrate that current frontier LLMs lack reliable native sampling under standard decoding, but do not constitute an impossibility proof for future architectures or training paradigms. Although we benchmark 15 canonical distributions across multiple complexity tiers, all targets are explicitly specified; real-world stochastic processes may involve implicit or context-dependent distributions beyond our experimental scope. Finally, the downstream tasks are used as controlled tests with explicit distributional constraints, to show that sampling failures alone can induce generation-stage bias, rather than to provide a comprehensive fairness analysis.

\section{Ethical Considerations}
To our knowledge, this work is the first to systematically demonstrate how native sampling failures propagate into downstream bias during the generation process. In the MCQ experiment, models' inability to follow uniform positional constraints directly produces answer-position bias, compromising the fairness of LLM-generated evaluation materials. In the attribute-constrained generation experiment, models' failure to adhere to demographic distributions causes sampling-induced bias to be embedded directly into synthetic data. These findings carry significant implications for high-stakes applications. When LLMs are deployed for social simulation, synthetic data generation, or randomized decision-making, their outputs are often implicitly treated as valid probabilistic samples. Our results demonstrate that this assumption is fundamentally unwarranted. We urge the community to critically examine the potential consequences of sampling infidelity in application contexts where distributional accuracy is essential for fairness, validity, or safety.

\bibliography{custom}

\appendix

\section{Decoding Parameter Ablations}
\label{app:Decoding}

Table~\ref{tab:temp_ablation_combined} and \ref{tab:topp_ablation_combined}  present the results of our decoding ablations (GPT-4o and Gemma; all 15 distributions; Batch + Independent; $N{=}1000$) over temperature $\in \{0.2, 0.5, 1.0, 1.2\}$ and \texttt{top-p} $\in \{0.9, 0.95, 1.0\}$. We did not include \texttt{top\_k} in the cross-model ablation because \texttt{top-k} is not exposed consistently across all APIs like GPT-4o. The overall conclusion remains unchanged: decoding sweeps do not consistently recover faithful native sampling, and the Independent protocol remains broadly poor across settings, which strengthens our main claim that LLMs lack a functional internal sampler.

\begin{table*}[t]
\centering
\scriptsize
\setlength{\tabcolsep}{4pt}
\begin{tabular}{l|cccc|cccc}
\toprule
\textbf{Distribution}
& \multicolumn{4}{c|}{\textbf{GPT-4o}}
& \multicolumn{4}{c}{\textbf{Gemma-3}} \\
& $T=0.2$ & $T=0.5$ & $T=1.0$ & $T=1.2$
& $T=0.2$ & $T=0.5$ & $T=1.0$ & $T=1.2$ \\
\midrule
\multicolumn{9}{l}{\textit{Batch Mode}} \\
\multicolumn{9}{l}{\textit{Discrete Distributions}} \\
Bernoulli & 0.28 & 0.28 & 4e-05$^*$ & 0.11 & 0.03 & 0.08 & 0.06 & 0.09 \\
Binomial  & 1.3  & 1.3  & 0.26       & 0.44 & 0.17 & 1.3  & 1.1  & 1.3 \\
Poisson   & 1.2  & 1.7  & 0.26       & 0.78 & 1.0  & 1.8  & 1.0  & 0.77 \\
\midrule
\multicolumn{9}{l}{\textit{Continuous Distributions}} \\
Uniform     & 0.04 & 0.05 & 0.02$^*$ & 0.01$^*$ & 0.03 & 0.07 & 0.03$^*$ & 0.02$^*$ \\
Gaussian    & 0.16 & 0.23 & 0.10$^*$ & 1.9$^*$  & 0.16 & 0.15 & 0.15      & 0.10$^*$ \\
Beta        & 0.17 & 0.10 & 0.10      & 0.02$^*$ & 0.06 & 0.06 & 0.06      & 0.05 \\
Exponential & 1.0  & 1.0  & 0.24      & --       & 0.34 & 0.37 & 0.38      & 0.37 \\
Cauchy      & 6.1  & 6.2  & 3.3$^*$   & 4.3$^*$  & 11.9 & 5.7  & 6.6       & 5.5 \\
$ t $       & 0.33 & 0.41 & 0.13$^*$  & --       & 0.38 & 5.8  & 0.48      & 0.32$^*$ \\
$\chi^2$    & 1.6  & 2.4  & 0.93      & 1.2      & 11.7 & 0.79 & 1.6       & 0.57$^*$ \\
$F$         & 0.78 & 0.59 & 0.45      & 4.9      & 16.8 & 2.7  & 0.98      & 6.0 \\
Gamma       & 1.8  & 2.1  & 0.78      & 1.2      & 1.2  & 6.7  & 2.0       & 3.2 \\
Weibull     & 0.88 & 0.29 & 0.32      & 1.6      & 0.31 & 0.37 & 0.13      & 3.5 \\
Laplace     & 0.26 & 0.97 & 0.32$^*$  & 0.32     & 0.25 & 0.34 & 0.35      & 0.31 \\
Logistic    & 1.4  & 1.2  & 0.73      & --       & 6.5  & 0.70 & 0.40      & 3.9 \\
\midrule
\multicolumn{9}{l}{\textit{Independent Mode}} \\
\multicolumn{9}{l}{\textit{Discrete Distributions}} \\
Bernoulli & 0.32 & 0.31 & 0.31 & 0.21 & 0.32 & 0.32 & 0.32 & 0.32 \\
Binomial  & 1.1  & 0.84 & 0.83 & 0.71 & 1.5  & 1.4  & 1.4  & 0.75 \\
Poisson   & 1.1  & 0.75 & 0.52 & 0.44 & 3.3  & 3.3  & 2.5  & 3.2 \\
\midrule
\multicolumn{9}{l}{\textit{Continuous Distributions}} \\
Uniform     & 0.18 & 0.17 & 0.16 & 0.16 & 0.21 & 0.17 & 0.17 & 0.14 \\
Gaussian    & 0.60 & 0.44 & 0.27 & 0.25 & 0.81 & 0.81 & 0.72 & 0.75 \\
Beta        & 0.12 & 0.10 & 0.08 & 0.08 & 0.21 & 0.21 & 0.15 & 0.13 \\
Exponential & 0.62 & 0.57 & 0.27 & 0.45 & 0.68 & 0.62 & 0.49 & 0.46 \\
Cauchy      & 3.7  & 3.4  & 2.8  & 2.5$^*$ & 3.8 & 3.7 & 3.6 & 3.7 \\
$ t $       & 0.99 & 0.53 & 0.40 & 0.59 & 1.3 & 1.3 & 1.4 & 1.1 \\
$\chi^2$    & 2.2  & 1.6  & 1.1  & 1.2  & 2.5 & 2.1 & 2.9 & 1.9 \\
$F$         & 1.3  & 1.3  & 0.56 & 1.3  & 3.7 & 3.4 & 1.2 & 3.0 \\
Gamma       & 2.0  & 2.0  & 1.2  & 1.8  & 2.1 & 2.0 & 1.8 & 1.8 \\
Weibull     & 0.42 & 0.39 & 0.30 & 0.33 & 0.75 & 0.75 & 0.47 & 0.68 \\
Laplace     & 0.97 & 0.60 & 0.36 & 0.25 & 1.1 & 1.0 & 0.64 & 0.86 \\
Logistic    & 1.2  & 1.0  & 0.66 & 0.68 & 1.3 & 1.3 & 1.2 & 1.2 \\
\bottomrule
\end{tabular}
\caption{\textbf{Temperature Ablation.} Each cell reports Wasserstein distance $\mathcal{W}_1$; $^*$ indicates passing the corresponding statistical test ($p>0.01$). For $T=1.2$, GPT-4o produced garbled and unparseable outputs for the Exponential, $t$-distribution, and Logistic settings in five repeated attempts in batch mode; these cases are therefore reported as ``--''.}
\label{tab:temp_ablation_combined}
\end{table*}

\begin{table*}[t]
\centering
\scriptsize
\setlength{\tabcolsep}{5pt}
\begin{tabular}{l|ccc|ccc}
\toprule
\textbf{Distribution}
& \multicolumn{3}{c|}{\textbf{GPT-4o}}
& \multicolumn{3}{c}{\textbf{Gemma-3}} \\
& $p=0.9$ & $p=0.95$ & $p=1.0$
& $p=0.9$ & $p=0.95$ & $p=1.0$ \\
\midrule
\multicolumn{7}{l}{\textit{Batch Mode}} \\
\multicolumn{7}{l}{\textit{Discrete Distributions}} \\
Bernoulli & 0.10 & 6e-03$^*$ & 4e-05$^*$ & 0.12 & 0.19 & 0.06 \\
Binomial  & 0.91 & 0.85       & 0.26       & 1.0  & 1.4  & 1.1 \\
Poisson   & 0.68 & 0.38       & 0.26       & 0.90 & 1.7  & 1.0 \\
\midrule
\multicolumn{7}{l}{\textit{Continuous Distributions}} \\
Uniform     & 0.09 & 0.02$^*$ & 0.02$^*$ & 0.01$^*$ & 0.10 & 0.03$^*$ \\
Gaussian    & 0.31 & 0.20      & 0.10$^*$ & 0.18      & 0.14 & 0.15 \\
Beta        & 0.17 & 0.13      & 0.10      & 0.05      & 0.03 & 0.06 \\
Exponential & 1.0  & 0.50      & 0.24      & 0.60      & 0.37 & 0.38 \\
Cauchy      & 5.8  & 5.8       & 3.3$^*$   & 5.6       & 5.6  & 6.6 \\
$ t $       & 0.82 & 0.44      & 0.13$^*$  & 0.34      & 2.7  & 0.48 \\
$\chi^2$    & 1.9  & 3.1       & 0.93      & 0.97      & 0.64 & 1.6 \\
$F$         & 0.47 & 0.31      & 0.45      & 0.47      & 0.30 & 0.98 \\
Gamma       & 1.5  & 1.2       & 0.78      & 2.0       & 1.1  & 2.0 \\
Weibull     & 0.46 & 0.25      & 0.32      & 0.29      & 0.31 & 0.13 \\
Laplace     & 0.44 & 0.38      & 0.32$^*$  & 0.32      & 4.0  & 0.35 \\
Logistic    & 0.92 & 0.79      & 0.73      & 0.71      & 0.97 & 0.40 \\
\midrule
\multicolumn{7}{l}{\textit{Independent Mode}} \\
\multicolumn{7}{l}{\textit{Discrete Distributions}} \\
Bernoulli & 0.31 & 0.26 & 0.31 & 0.32 & 0.32 & 0.32 \\
Binomial  & 0.77 & 0.88 & 0.83 & 0.72 & 0.70 & 1.4 \\
Poisson   & 0.71 & 0.58 & 0.52 & 3.3  & 3.2  & 2.5 \\
\midrule
\multicolumn{7}{l}{\textit{Continuous Distributions}} \\
Uniform     & 0.17 & 0.17 & 0.16 & 0.14 & 0.13 & 0.17 \\
Gaussian    & 0.31 & 0.28 & 0.27 & 0.81 & 0.79 & 0.72 \\
Beta        & 0.09 & 0.09 & 0.08 & 0.16 & 0.19 & 0.15 \\
Exponential & 0.55 & 0.54 & 0.27 & 0.52 & 0.47 & 0.49 \\
Cauchy      & 3.1  & 2.9  & 2.8  & 3.7  & 3.7  & 3.6 \\
$ t $       & 0.55 & 0.60 & 0.40 & 1.3  & 1.2  & 1.4 \\
$\chi^2$    & 1.4  & 1.3  & 1.1  & 1.9  & 2.0  & 2.9 \\
$F$         & 1.3  & 1.2  & 0.56 & 3.2  & 3.1  & 1.2 \\
Gamma       & 1.9  & 1.9  & 1.2  & 2.0  & 1.9  & 1.8 \\
Weibull     & 0.38 & 0.36 & 0.30 & 0.73 & 0.73 & 0.47 \\
Laplace     & 0.36 & 0.38 & 0.36 & 0.91 & 0.87 & 0.64 \\
Logistic    & 0.87 & 0.83 & 0.66 & 1.2  & 1.2  & 1.2 \\
\bottomrule
\end{tabular}
\caption{\textbf{Top-$p$ Ablation.} Each cell reports Wasserstein distance $\mathcal{W}_1$; $^*$ indicates passing the corresponding statistical test ($p>0.01$).}
\label{tab:topp_ablation_combined}
\end{table*}

\section{Multivariate Extension}
\label{app:multivariate_extension}

We further extend our evaluation to a bivariate Gaussian setting. Specifically, we consider a zero-mean Gaussian with correlation $\rho=0.5$:
\[
(X,Y) \sim \mathcal{N}(\mathbf{0}, \Sigma), \qquad
\Sigma =
\begin{pmatrix}
1 & 0.5 \\
0.5 & 1
\end{pmatrix}.
\]
In addition to the two marginals, we evaluate the conditional distribution $Y \mid X > 0$. We test three models (GPT-4o, Kimi-K2, and Gemma-3-27B) under both batch and independent settings with $N{=}1000$. To assess multivariate fidelity, we measure: (i) marginal two-sample KS and $\mathcal{W}_1$ for $X$ and $Y$; and (ii) conditional two-sample KS and $\mathcal{W}_1$ for $Y \mid X > 0$. As shown in Table~\ref{tab:bivariate_gaussian_w1}, the multivariate extension reveals the same qualitative failure mode as our main experiments. In batch mode, all three models pass the KS test for the \(Y\) marginal, yet still fail to recover the conditional distribution \(Y \mid X > 0\). Under independent requests, fidelity degrades further across both marginals and the conditional distribution.

\begin{table}[t]
\centering
\scriptsize
\setlength{\tabcolsep}{7pt}
\begin{tabular}{llccc}
\toprule
\textbf{Model} & \textbf{Mode} & \textbf{$X$} & \textbf{$Y$} & \textbf{$Y\mid X>0$} \\
\midrule
GPT-4o      & Batch       & 1.4486 & 0.0496$^*$ & 0.2873 \\
Kimi-K2     & Batch       & 0.3115 & 0.2306$^*$ & 0.5258 \\
Gemma-3-27B & Batch       & 0.3395 & 0.1812$^*$ & 0.3994 \\
GPT-4o      & Independent & 0.4008 & 0.4480     & 0.2969 \\
Kimi-K2     & Independent & 0.3354 & 0.6079     & 0.2934 \\
Gemma-3-27B & Independent & 0.9125 & 0.6835     & 0.7576 \\
\bottomrule
\end{tabular}
\caption{\textbf{Bivariate Gaussian Extension ($\rho=0.5$, $N{=}1000$).}
Each cell reports Wasserstein distance $\mathcal{W}_1$. Columns $X$ and $Y$ denote marginal results, and $Y\mid X>0$ denotes the conditional result. $^*$ indicates that the corresponding two-sample KS test passes ($p>0.01$).}
\label{tab:bivariate_gaussian_w1}
\end{table}

\section{Detailed Sampling Fidelity Results}
As shown in Section~\ref{sec:result}, Figure \ref{fig:tier_comparison} illustrates the degradation of sampling fidelity as distributional complexity increases. Table~\ref{tab:appendix_tier} provides the comprehensive numerical breakdown of this trend, detailing Pass Rates and Wasserstein-1 ($\mathcal{W}_1$) distances for all 11 models across the three complexity tiers under both Batch and Independent protocols. The data confirms a monotonic degradation in performance as complexity increases.

\begin{table*}[t]
\centering
\small
\setlength{\tabcolsep}{3pt}
\begin{tabular}{l|cc|cc|cc|cc||cc|cc|cc|cc}
\toprule
& \multicolumn{8}{c||}{\textbf{Batch Mode}} & \multicolumn{8}{c}{\textbf{Independent Mode}} \\
& \multicolumn{2}{c|}{Tier I (3)} & \multicolumn{2}{c|}{Tier II (4)} & \multicolumn{2}{c|}{Tier III (8)} & \multicolumn{2}{c||}{Overall (15)}
& \multicolumn{2}{c|}{Tier I (3)} & \multicolumn{2}{c|}{Tier II (4)} & \multicolumn{2}{c|}{Tier III (8)} & \multicolumn{2}{c}{Overall (15)} \\
\textbf{Model} & Rate & $\mathcal{W}_1$ & Rate & $\mathcal{W}_1$ & Rate & $\mathcal{W}_1$ & Rate & $\mathcal{W}_1$
& Rate & $\mathcal{W}_1$ & Rate & $\mathcal{W}_1$ & Rate & $\mathcal{W}_1$ & Rate & $\mathcal{W}_1$ \\
\midrule
\multicolumn{17}{l}{\textit{Proprietary Models}} \\
GPT-5.2 & \textbf{67\%} & 0.08 & 0\% & 0.44 & 0\% & 1.14 & \textbf{13\%} & 0.74 & 0\% & 0.35 & 0\% & 0.76 & 0\% & 1.11 & 0\% & 0.86 \\
Gemini-3 & \textbf{33\%} & 0.07 & 0\% & 0.17 & \textbf{12\%} & 1.05 & \textbf{13\%} & 0.62 & 0\% & 0.41 & 0\% & 0.60 & 0\% & 1.20 & 0\% & 0.88 \\
GPT-4o & \textbf{100\%} & 0.04 & 0\% & 0.22 & \textbf{38\%} & 0.87 & \textbf{40\%} & 0.53 & 0\% & 0.24 & 0\% & 0.43 & 0\% & 0.92 & 0\% & 0.65 \\
\midrule
\multicolumn{17}{l}{\textit{Open-Weights Models}} \\
DeepSeek-V3.2 & \textbf{33\%} & 0.15 & 0\% & 0.19 & 0\% & 1.23 & \textbf{7\%} & 0.74 & 0\% & 0.25 & 0\% & 0.62 & 0\% & 4.17 & 0\% & 2.44 \\
Qwen3 & 0\% & 0.15 & 0\% & 0.68 & 0\% & 1.62 & 0\% & 1.08 & 0\% & 0.54 & 0\% & 2.24 & 0\% & 2.04 & 0\% & 1.79 \\
Gemma-3 & \textbf{33\%} & 0.08 & 0\% & 0.65 & 0\% & 1.56 & \textbf{7\%} & 1.02 & 0\% & 0.40 & 0\% & 1.14 & 0\% & 1.66 & 0\% & 1.27 \\
Mistral-3.2 & 0\% & 0.18 & 0\% & 0.60 & 0\% & 1.85 & 0\% & 1.18 & 0\% & 0.28 & 0\% & 0.76 & 0\% & 1.10 & 0\% & 0.85 \\
Kimi-K2 & \textbf{33\%} & 0.08 & \textbf{25\%} & 0.50 & \textbf{12\%} & 1.42 & \textbf{20\%} & 0.90 & 0\% & 0.30 & 0\% & 0.54 & 0\% & 0.99 & 0\% & 0.73 \\
Llama-3.3 & 0\% & 0.20 & 0\% & 0.31 & 0\% & 1.66 & 0\% & 1.01 & 0\% & 0.36 & 0\% & 1.10 & 0\% & 1.52 & 0\% & 1.18 \\
Llama-4 & 0\% & 0.20 & 0\% & 0.41 & \textbf{12\%} & 1.40 & \textbf{7\%} & 0.89 & \textbf{33\%} & 0.38 & 0\% & 0.70 & 0\% & 1.64 & \textbf{7\%} & 1.14 \\
GPT-OSS & \textbf{67\%} & 0.09 & 0\% & 0.45 & 0\% & 1.41 & \textbf{13\%} & 0.89 & 0\% & 0.34 & 0\% & 0.78 & 0\% & 1.05 & 0\% & 0.84 \\
\bottomrule
\end{tabular}
\caption{\textbf{Main Results: Fidelity of Native Sampling.} 
Pass Rate (\%) and mean $\mathcal{W}_1$ by complexity tier.
For discrete distributions: $\chi^2$ GoF Test; for continuous: Two-Sample KS Test ($\alpha=0.01$).
Lower $\mathcal{W}_1$ indicates better distributional fit.}
\label{tab:appendix_tier}
\end{table*}

\section{Significance-Threshold Sensitivity}
\label{app:alpha_sensitivity}

To assess sensitivity to the choice of significance threshold, we additionally report pass rates under $\alpha=0.05$ and $\alpha=0.005$, alongside our main setting $\alpha=0.01$. As shown in Table~\ref{tab:alpha_sensitivity}, the results under $\alpha=0.005$ remain broadly similar to those under $\alpha=0.01$, and the qualitative trends as well as our main conclusions remain unchanged. This indicates that our findings are not an artifact of a particular threshold choice.

\begin{table*}[t]
\centering
\small
\setlength{\tabcolsep}{4pt}
\begin{tabular}{l|ccc|ccc}
\toprule
\textbf{Model}
& \multicolumn{3}{c|}{\textbf{Batch}}
& \multicolumn{3}{c}{\textbf{Independent}} \\
& $\alpha=0.05$ & $\alpha=0.01$ & $\alpha=0.005$
& $\alpha=0.05$ & $\alpha=0.01$ & $\alpha=0.005$ \\
\midrule
GPT-4o   & 33\% & 40\% & 40\% & 0\% & 0\% & 0\% \\
Kimi     & 13\% & 20\% & 33\% & 0\% & 0\% & 0\% \\
Gemini-3 & 13\% & 13\% & 13\% & 0\% & 0\% & 0\% \\
GPT-5.2  & 7\%  & 13\% & 13\% & 0\% & 0\% & 0\% \\
GPT-OSS  & 7\%  & 13\% & 13\% & 0\% & 0\% & 0\% \\
DeepSeek & 7\%  & 7\%  & 7\%  & 0\% & 0\% & 0\% \\
Llama-4  & 0\%  & 7\%  & 7\%  & 7\% & 7\% & 7\% \\
Gemma-3  & 0\%  & 7\%  & 7\%  & 0\% & 0\% & 0\% \\
Qwen3    & 0\%  & 0\%  & 0\%  & 0\% & 0\% & 0\% \\
Mistral  & 0\%  & 0\%  & 0\%  & 0\% & 0\% & 0\% \\
Llama-3.3& 0\%  & 0\%  & 0\%  & 0\% & 0\% & 0\% \\
\bottomrule
\end{tabular}
\caption{\textbf{Pass-rate sensitivity to the significance threshold.}
Each entry reports the percentage of distributions passed under the corresponding significance level $\alpha$.}
\label{tab:alpha_sensitivity}
\end{table*}

\section{Distribution-Sampling Prompt Example}
\label{app:main_prompt}

To ensure reproducibility, we provide a representative example of the original batch prompt used in our distribution-sampling experiments. The prompt below shows the exact instruction style used to ask the model to generate $N$ samples from a target distribution. Since prior studies suggest that LLMs do not reliably follow exact counting or length constraints \citep{yuan-etal-2025-following, xu2025llmgeniusparadoxlinguistic}, we incorporated explicit instruction-level requirements into the prompt to encourage adherence to the requested number of samples and the prescribed output format.

\begin{table*}[!htb]
\begin{tcolorbox}[
  title=Original Batch Prompt Example (Uniform Distribution),
  colback=white,
  colframe=black!60
]
\cotinstr{\textbf{User:}\\
You are a random number generator. Your task is to generate exactly \{n\_samples\} independent samples from a Uniform distribution on the interval [\{low\}, \{high\}].\medskip\\

\textbf{INSTRUCTIONS:}\\
1. You MUST output EXACTLY \{n\_samples\} numbers. Not \{n\_samples\}$-1$, not \{n\_samples\}$+1$, not any other quantity. And if the count is wrong, your output is considered invalid.\\
2. Each number must be between \{low\} and \{high\} (inclusive).\\
3. Numbers should be randomly distributed across [\{low\}, \{high\}].\\
4. Output ONLY the numbers, separated by commas or spaces.\\
5. Use decimal numbers with varying precision.\\
6. Do NOT add any explanations, labels, or text.\medskip\\

\textbf{Final instruction:}\\
Now generate EXACTLY \{n\_samples\} samples from $U(\{low\}, \{high\})$:
}
\end{tcolorbox}
\caption{Representative original batch prompt used for distribution sampling, shown here for the Uniform distribution. Other distributions follow the same overall instruction style, with distribution-specific descriptions and validity constraints instantiated accordingly.}
\label{tab:prompt-distribution-uniform}
\end{table*}

\section{Downstream Application Experimental Prompts}
\label{app:downstream_prompt}
To ensure reproducibility, we provide the exact system instructions and user prompts employed in our downstream application experiments in Table~\ref{tab:prompt-mcq} and Table~\ref{tab:prompt-joint}.

% Requires: \usepackage{tcolorbox}
% Define commands if not already defined:
% \newcommand{\vanillainstr}[1]{\colorbox{gray!15}{#1}}
% \newcommand{\cotinstr}[1]{#1}

\begin{table*}[!htb]
\begin{tcolorbox}[title=MCQ Answer Distribution (Medical Multiple-Choice Generation),
  colback=white, colframe=black!60]
\vanillainstr{\textbf{System:} You are a medical education expert who creates high-quality multiple-choice questions for medical students and professionals.}\medskip\\
\cotinstr{\textbf{User:}\\
Please generate a medical multiple-choice question (single answer, 4 options). The question should cover medical knowledge and be of moderate difficulty.\medskip\\
Please strictly follow this format:\medskip\\
\hspace*{1em}Question: [Question content]\\
\hspace*{1em}A. [Option A content]\\
\hspace*{1em}B. [Option B content]\\
\hspace*{1em}C. [Option C content]\\
\hspace*{1em}D. [Option D content]\\
\hspace*{1em}Correct Answer: [A/B/C/D]\\
\hspace*{1em}Explanation: [Brief explanation]\medskip\\
Requirements:\\
\hspace*{1em}(1) The question should have practical medical value.\\
\hspace*{1em}(2) All four options should be plausible with reasonable distractors.\\
\hspace*{1em}(3) Only one correct answer.\\
\hspace*{1em}(4) Output directly without any additional content.\\
\hspace*{1em}(5) Cover different medical knowledge areas (e.g., internal medicine, surgery, pharmacology, pathology, diagnostics).\\
\hspace*{1em}(6) The correct answer should be evenly distributed among A, B, C, D options to avoid bias toward any particular option.}
\end{tcolorbox}
\caption{Prompt for MCQ Answer Distribution experiment. The LLM is instructed to generate medical questions with uniformly distributed correct answers (25\% each for A, B, C, D).}
\label{tab:prompt-mcq}
\end{table*}

% Requires: \usepackage{tcolorbox}
% Define commands if not already defined:
% \newcommand{\vanillainstr}[1]{\colorbox{gray!15}{#1}}
% \newcommand{\cotinstr}[1]{#1}

\begin{table*}[!htb]
\begin{tcolorbox}[title=Joint Attribute Sampling (Text-to-Image Prompt Generation),
  colback=white, colframe=black!60]
\vanillainstr{\textbf{System:} You are a prompt generator that follows statistical distributions precisely.}\medskip\\
\cotinstr{\textbf{User:}\\
You are generating text-to-image prompts that must SIMULTANEOUSLY follow multiple statistical distributions.\medskip\\
\textbf{=== TARGET DISTRIBUTIONS ===}\medskip\\
\textbf{1. GENDER} (U.S. Census Bureau 2024):\\
\hspace*{2em}Male: 49.49\% \hspace{1em} Female: 50.51\%\medskip\\
\textbf{2. RACE/ETHNICITY} (U.S. Census Bureau 2024, NC-EST2024-SR11H):\\
\hspace*{2em}White (Non-Hispanic): 57.46\%\\
\hspace*{2em}Hispanic/Latino: 20.02\%\\
\hspace*{2em}Black (Non-Hispanic): 12.63\%\\
\hspace*{2em}Asian (Non-Hispanic): 6.49\%\\
\hspace*{2em}Others (AIAN, NHPI, Mixed): 3.40\%\medskip\\
\textbf{3. HEIGHT} (Normal Distribution):\\
\hspace*{2em}$\mathcal{N}(169.0, 10.0^2)$ cm\\
\hspace*{2em}$\sim$68\% should be between 159--179 cm\\
\hspace*{2em}$\sim$95\% should be between 149--189 cm\medskip\\
\textbf{4. COAT COLOR} (Uniform Distribution):\\
\hspace*{2em}7 colors with EQUAL probability (14.29\% each):\\
\hspace*{2em}Black, White, Red, Blue, Green, Yellow, Brown\medskip\\
\textbf{=== YOUR TASK ===}\\
Generate ONE text-to-image prompt describing a person wearing a coat.\\
You must INDEPENDENTLY sample each attribute according to its distribution above.\medskip\\
\textbf{=== OUTPUT FORMAT (STRICTLY FOLLOW) ===}\\
\hspace*{1em}[Gender] [Race] [Height in cm] [Coat Color]\\
\hspace*{1em}Prompt: <your creative prompt>\medskip\\
\textbf{=== EXAMPLE ===}\\
\hspace*{1em}[Female] [Hispanic] [165] [Blue]\\
\hspace*{1em}Prompt: A Hispanic woman, 165cm tall, wearing a blue wool coat, walking through a sunlit autumn park.\medskip\\
\textbf{=== GENERATE ONE PROMPT NOW ===}\\
Remember to sample EACH attribute independently according to its target distribution.}
\end{tcolorbox}
\caption{Prompt for Joint Attribute Sampling experiment. The LLM must simultaneously sample four attributes (Gender, Race, Height, Color) according to their respective target distributions.}
\label{tab:prompt-joint}
\end{table*}

\section{Fine-Grained Attribute Analysis}

To isolate each model's capability in distribution-constrained generation, we conducted four independent attribute sampling experiments testing Gender, Race/Ethnicity, Height, and Coat Color separately. For each attribute, we prompted five state-of-the-art LLMs to generate $N=1000$ independent samples following explicitly specified target distributions. Unlike the joint experiment where models must simultaneously control multiple attributes, these independent tests measure single-attribute adherence in isolation. The target distributions are identical to those used in the joint experiment. We provided explicit distribution constraints and sampling instructions in each prompt to ensure models were fully aware of the target probabilities. All experiments used a temperature $T=1.0$ with default nucleus sampling parameters. Statistical significance was assessed using $\chi^2$ goodness-of-fit tests for categorical attributes (Gender, Race, Color) and the Kolmogorov-Smirnov test for the continuous attribute (Height), with significance threshold $\alpha=0.01$.

Tables~\ref{tab:ind_gender} through \ref{tab:ind_color} present the complete results, further confirming systematic failures in distribution-constrained generation across all models and attributes. Only 1 out of 20 experiments (DeepSeek on Gender, $\chi^2=2.3$, $p=0.127$) passed statistical testing ($\alpha=0.01$). The independent experiments revealed three critical failure patterns: (1) Demographic bias in Gender and Race sampling, where models exhibited extreme skews  (2) Variance collapse in Height generation, where all models achieved only 7--67\% of the target standard deviation ($\sigma=10.0$ cm), with Llama-4 collapsing to $\sigma=0.7$ cm, and (3) Catastrophic mode collapse in Color sampling, where models concentrated $>75$\% of outputs on 1--2 colors despite explicit uniform distribution instructions. These findings indicate that current LLMs fundamentally struggle with stochastic generation: even with explicit distribution constraints, models fail to achieve statistically valid random sampling, with important implications for downstream applications.

\begin{table*}[h]
\centering
\begin{tabular}{@{}lccccc|c@{}}
\toprule
& {Llama-3.3} & {DeepSeek} & {Qwen3} & {Llama-4} & {GPT-OSS} & {Target} \\
\midrule
Male (\%) & 3.1 & 51.9 & 2.3 & 0.0 & 0.2 & 49.5 \\
Female (\%) & 96.9 & 48.1 & 97.7 & 100.0 & 99.8 & 50.5 \\
$\chi^2$ & 860.9 & 2.3 & 890.9 & 979.8 & 971.9 & -- \\
\bottomrule
\end{tabular}
\caption{Independent Experiment: Gender Distribution (N=1000)}
\label{tab:ind_gender}
\end{table*}

\begin{table*}[h]
\centering
\begin{tabular}{@{}lccccc|c@{}}
\toprule
& {Llama-3.3} & {DeepSeek} & {Qwen3} & {Llama-4} & {GPT-OSS} & {Target} \\
\midrule
White (\%) & 95.5 & 43.3 & 44.5 & 93.6 & 71.1 & 57.5 \\
Hispanic (\%) & 0.0 & 14.0 & 26.1 & 6.0 & 13.4 & 20.0 \\
Black (\%) & 1.9 & 26.3 & 26.6 & 0.3 & 9.9 & 12.6 \\
Asian (\%) & 2.6 & 13.2 & 1.7 & 0.1 & 5.6 & 6.5 \\
Other (\%) & 0.0 & 3.2 & 1.1 & 0.0 & 0.0 & 3.4 \\
$\chi^2$ & 600.5 & 270.5 & 253.1 & 542.8 & 95.4 & -- \\
\bottomrule
\end{tabular}
\caption{Independent Experiment: Race/Ethnicity Distribution (N=1000)}
\label{tab:ind_race}
\end{table*}

\begin{table*}[h]
\centering
\begin{tabular}{@{}lccccc|c@{}}
\toprule
& {Llama-3.3} & {DeepSeek} & {Qwen3} & {Llama-4} & {GPT-OSS} & {Target} \\
\midrule
$\mu$ (cm) & 171.1 & 169.7 & 169.5 & 173.3 & 170.5 & 169.0 \\
$\sigma$ (cm) & 4.5 & 5.4 & 1.4 & 0.7 & 6.7 & 10.0 \\
KS stat & 0.33 & 0.21 & 0.50 & 0.66 & 0.25 & -- \\
\bottomrule
\end{tabular}
\caption{Independent Experiment: Height Distribution (N=1000, Target: $\mathcal{N}(169.0, 10.0^2)$ cm)}
\label{tab:ind_height}
\end{table*}

\begin{table*}[h]
\centering
\begin{tabular}{@{}lccccc|c@{}}
\toprule
& {Llama-3.3} & {DeepSeek} & {Qwen3} & {Llama-4} & {GPT-OSS} & {Target} \\
\midrule
Black (\%) & 0.0 & 2.0 & 15.2 & 0.0 & 0.0 & 14.3 \\
White (\%) & 0.0 & 6.2 & 3.1 & 0.0 & 0.1 & 14.3 \\
Red (\%) & 0.0 & 15.2 & 9.9 & 0.0 & 18.9 & 14.3 \\
Blue (\%) & 0.0 & 27.8 & 53.8 & 0.0 & 3.6 & 14.3 \\
Green (\%) & 98.4 & 26.1 & 11.6 & 75.8 & 65.5 & 14.3 \\
Yellow (\%) & 1.6 & 17.2 & 4.2 & 24.2 & 11.9 & 14.3 \\
Brown (\%) & 0.0 & 5.5 & 2.2 & 0.0 & 0.0 & 14.3 \\
$\chi^2$ & 5779.6 & 437.5 & 1373.1 & 3431.9 & 2361.4 & -- \\
\bottomrule
\end{tabular}
\caption{Independent Experiment: Coat Color Distribution (N=1000, Target: Uniform 14.3\% each)}
\label{tab:ind_color}
\end{table*}

\section{The Use of Large Language Models (LLMs)}
\label{app:llm-usage}
LLM is used only to aid writing quality (proofreading and polishing grammar). No ideas, claims, methods, results, or references are generated by LLMs. All content decisions and revisions are made by the authors.

\end{document}